\documentclass{article}
\usepackage{spconf,amsmath,epsfig}
\usepackage[colorlinks,
            linkcolor=red,
            citecolor=green,
            urlcolor=blue]{hyperref}

\usepackage{microtype}
\usepackage{amsfonts}
\usepackage{graphicx}
\usepackage{color, colortbl}
\usepackage[disable]{todonotes}
\usepackage{calc}  
\usepackage{array}
\usepackage{enumitem}
\usepackage{tabularx}
\usepackage{etoolbox}
\usepackage{caption} 
\usepackage{fancyhdr}

\usepackage{booktabs}
\usepackage{amssymb}
\usepackage [english]{babel}
\usepackage [autostyle, english = american]{csquotes}
\MakeOuterQuote{"}

\usepackage{xspace}

\makeatletter
\DeclareRobustCommand\onedot{\futurelet\@let@token\@onedot}
\def\@onedot{\ifx\@let@token.\else.\null\fi\xspace}

\def\eg{\emph{e.g}\onedot} 
\def\ie{\emph{i.e}\onedot}

\newcommand{\owner}[1]{
\empty
\relax
}

\usepackage{cleveref}

\usepackage{multicol}

\crefname{subfigure}{subfigure}{subfigures}
\Crefname{subfigure}{Subfigure}{Subfigures}
\crefname{subfigure}{}{}
\Crefname{subfigure}{}{}
\creflabelformat{subfigure}{(#2#1#3)}
\crefmultiformat{subfigure}{(#2#1#3}{ and~#2#1#3)}{, #2#1#3}{ and~#2#1#3)}
\crefname{appsec}{appendix}{appendices}
\Crefname{appsec}{Appendix}{Appendices}

\usepackage{makecell}

\makeatletter
\let\latex@@caption\caption
\NewDocumentCommand{\captext}{+o+m}{%
  \IfValueTF{#1}{%
      \latex@@caption[#1]{\textbf{#1}\quad{}\@ifnextchar\par\@gobble\relax#2}%
  }{%
    \latex@@caption[ERROR-NO-SHORT-TEXT]{#2}
  }
}
\makeatother

\newcommand{\caplbl}[1]{}

\usepackage{enumitem}
\setlist{nosep}

\usepackage{xurl}

\renewcommand{\paragraph}[1]{\noindent\textbf{#1}\quad}

\title{GeoWATCH for Detecting Heavy Construction in Heterogeneous Time Series of Satellite Images}
\oneauthor
{
Jon Crall, 
Connor Greenwell, 
David Joy, 
Matthew Leotta,
Aashish Chaudhary,
Anthony Hoogs
}
{Kitware, Inc.}

 \usepackage{fancyhdr}
 \fancypagestyle{firststyle}
 {
 \fancyhf{} 
 \fancyfoot[LO]{
 \scriptsize Copyright 2024 IEEE. Published in the 2024 IEEE International Geoscience and Remote Sensing Symposium (IGARSS 2024), scheduled for 7 - 12 July, 2024 in Athens, Greece. Personal use of this material is permitted. However, permission to reprint/republish this material for advertising or promotional purposes or for creating new collective works for resale or redistribution to servers or lists, or to reuse any copyrighted component of this work in other works, must be obtained from the IEEE. Contact: Manager, Copyrights and Permissions / IEEE Service Center / 445 Hoes Lane / P.O. Box 1331 / Piscataway, NJ 08855-1331, USA. Telephone: + Intl. 908-562-3966.}
 }

\begin{document}
%
\maketitle

\thispagestyle{firststyle}








\begin{abstract}




Learning from multiple sensors is challenging due to spatio-temporal misalignment and differences in resolution and captured spectra.
To that end, we introduce \textit{GeoWATCH}, a flexible framework for training models on long sequences of satellite images sourced from multiple sensor platforms, which is designed to handle image classification, activity recognition, object detection, or object tracking tasks.
Our system includes a novel partial weight loading mechanism based on sub-graph isomorphism which allows for continually training and modifying a network over many training cycles.
This has allowed us to train a lineage of models over a long period of time, which we have observed has improved performance as we adjust configurations while maintaining a core backbone.

\end{abstract}

\section{Introduction}
\label{sec:intro}


The deployment of remote sensing models in realistic settings features a number of challenges: sensor noise, full or partial occlusion from shadows and clouds, and data sparsity due to the fixed revisit rates of many sensor platforms.
Taken together, these challenges complicate long term Earth monitoring from a single sensor platform as high quality images are received less frequently.
A common strategy to increase the number of viable images is to incorporate readings from additional sensor platforms.
To overcome spatial, resolution, and spectral misalignment between multiple sensor platforms, harmonized products such as HLS \cite{claverie2018harmonized} have been introduced.
However, this limits training and inference to a fixed set of sensors.
In this paper we introduce \textit{GeoWATCH}, a software package for remote sensing computer vision tasks which is designed to handle multiple sensors, long time frames, be robust to missing and noisy data, and more.

While \textit{GeoWATCH} is designed to be a general use platform, it was initially built to solve the IARPA SMART large-scale heavy construction detection task.
The goal of the SMART program is to detect, characterize, and monitor 
anthropogenic or natural processes using a multiple sources of satellite imagery
collected over time.
The motivating use-case is the detection of heavy construction over a broad area of space and time.
Specifically, we construct a data cube of Landsat, Sentinel-2, and WorldView imagery covering
several large spatial regions over a period of almost eight years.
The goal is to predict a polygon on construction sites along with a start and end date.

\begin{figure}[t!]
    \centering
    \includegraphics[width=1\linewidth]{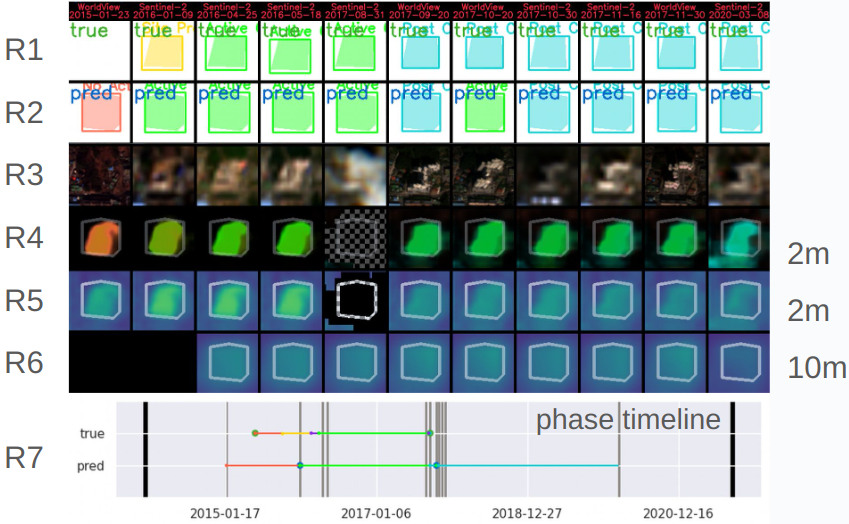}
    \captext[GeoWATCH Detection Example.]{


    Example prediction for a heavy construction site in the validation dataset. 
    Rows 1 and 2 display true and predicted polygons. 
    Row 3 presents the image data. 
    Rows 4 and 5 feature the 2m GSD phase and saliency heatmaps. 
    Row 6 displays the 10m GSD saliency heatmap. 
    Row 7 compares true and predicted timelines. 
    Category colors include red for "No Activity", yellow for "Site Preparation", green for "Active Construction", and blue for "Post Construction."
    }
    \label{fig:prediction_examle}
\end{figure}

Using GeoWATCH our algorithms are designed to make a prediction at every pixel in space time (\ie{} the output heatmaps are the same resolution as the input images) of 2 heads: ``saliency'' which indicate if any construction is happening, and ``class'', if it is one of 2 phases: ``Site Preparation'' or ``Active Construction''. We then extract polygons from these heatmaps and assign a start and end date. We operate in two stages: (1) broad area search for candidate sites at 10m GSD and (2) activity characterization and validation of candidate sites a 2m GSD. An example of a site detected with our system is shown in \Cref{fig:prediction_examle}.

Our contributions are:
1) An extension of MS-COCO~\cite{lin_microsoft_2014} called KWCoco, designed to be more suitable for geospatial data, and which provides the ability to sample large multispectral images at a virtual resolution using a novel "Video View". 
2) A new open source framework for training and predicting with AI models on geospatial data with support for continual learning regularization \cite{dohare_loss_2023}, which we call \textit{GeoWATCH}. 
3) A method for transferring part of a network to another network with similar structure by finding a maximum subtree embedding / isomorphism  \cite{feruglio_maximum_2003}, termed ``Partial Weight Loading''. 
We present preliminary observations of a phenomena that we call "instillation", where our models trained with one set of input features retain performance even after that input is removed.
and 4) We have publicly released our code\footnote{\url{https://gitlab.kitware.com/computer-vision/geowatch}} via GitLab and model weights
via IPFS\cite{benet_ipfs_2014}.

\section{Related Work}
\label{sec:related}

Our GeoWATCH framework is related to other libraries.
TorchGEO \cite{stewart_torchgeo_2022} - Defines specific dataloaders for individual standardized datasets, whereas our system opts to define a data interchange standard and use the same dataloader on all problems.  MMSegmentation \cite{mmsegmentation} - is a powerful semantic segmentation library, but cannot handle large images.
RasterVision \cite{rastervision} - Is another open source geospatial deep learning library with similar capabilities, but to the best of our knowledge it does not have the ability to produce native resolution batches on time sequences from multiple sensors.  

We also explore new techniques related to existing algorithms. 
Loss-of-plasticity \cite{dohare_loss_2023} is the observation that networks gradually degenerate while training with SGD, and that selectively re-initializing neurons can mitigate this. 
In our work we accomplish this re-initialization with partial-weight-loading, wherein the weights of a partially loaded network are reinitialized. 
Distillation \cite{hinton_distilling_2015, hu_teacher-student_2023} is a technique to transfer knowledge from one model to another. This is done by training a student network to match predictions of a teacher network. In this work we observe a new way to transfer knowledge, which we call ``instillation''. This involves a partial weight transfer of a larger network to a smaller network followed by fine-tuning. 



\section{Data Interchange}
\label{sec:approach:data}
\owner{Jon}

\begin{figure}
    \centering
    \includegraphics[width=1\linewidth]{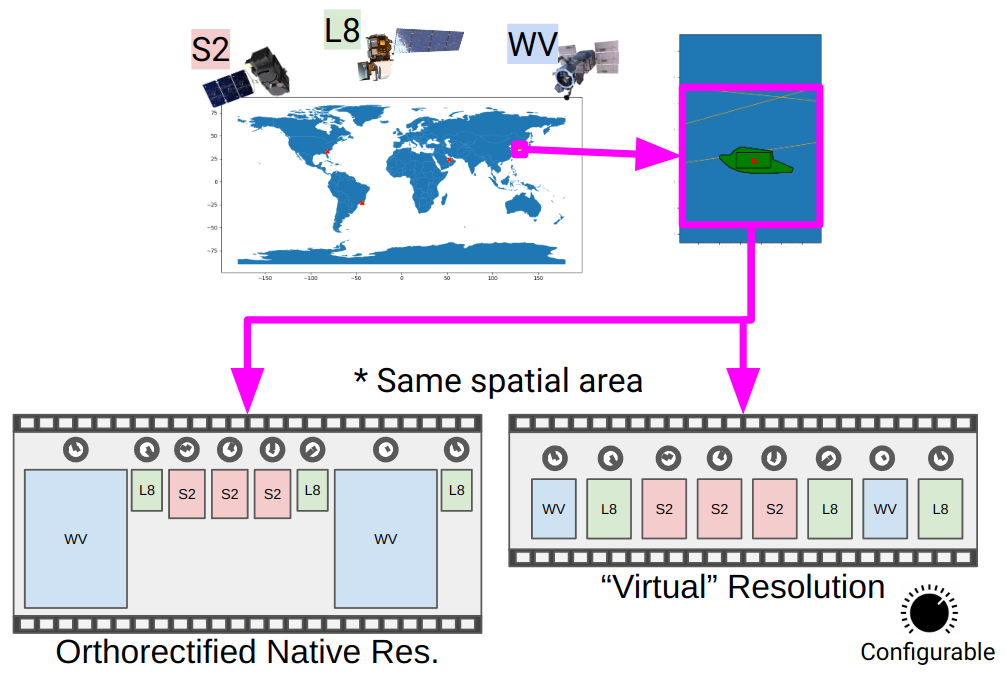}
    \captext[STAC-to-KWCoco.]{Given a region of interest, our system runs a STAC query and registers paths to the original images in a KWCoco file. The images are stored natively on disk and we can request heterogeneous subregions of spacetime at arbitrary resampled (or native) resolutions.}
    \label{fig:stac_to_KW-COCO}
\end{figure}

The main data interchange in our system for vision tasks is a KWCoco file, which is an extension of the well-known MS-COCO format~\cite{lin_microsoft_2014} that better handles a datacube of geospatial data. 
Specifically, COCO images are extended to allow for multi-spectral imagery stored across multiple assets, which could exist at different resolutions. 
Each COCO image can be registered as a frame in a video. 
Affine transformations are stored to warp between assets at different resolutions.

To prepare a region for training or prediction, our system runs STAC \cite{STAC} query and the results are indexed in a KWCoco file, which abstracts away geospatial information and allows computer vision algorithms to reference the data based on a virtual pixel space where all images area aligned. 
We call this virtual resolution a \textbf{video view}. 
This process is illustrated in \Cref{fig:stac_to_KW-COCO}.

Once raw GeoTIFFs are available on disk, and indexed by a KWCoco file, we can efficiently access them 
via the delayed image package \cite{delayed_image}. This tool utilizes the COG format \cite{COGFormat} to allow efficient access subregions of the cropped images (which might still be quite large --- \eg{} $4K \times 4K$ pixels is common), or coarser resolutions via overviews. It allows the developer to build a tree of image operations (\eg{} resampling / concatenation) and it is able to optimize this tree by fusing linear operations (\eg{} affine transforms) and replace downscales by 2 with an overview operation.

\begin{figure*}
    \centering
    \includegraphics[width=0.85\textwidth]{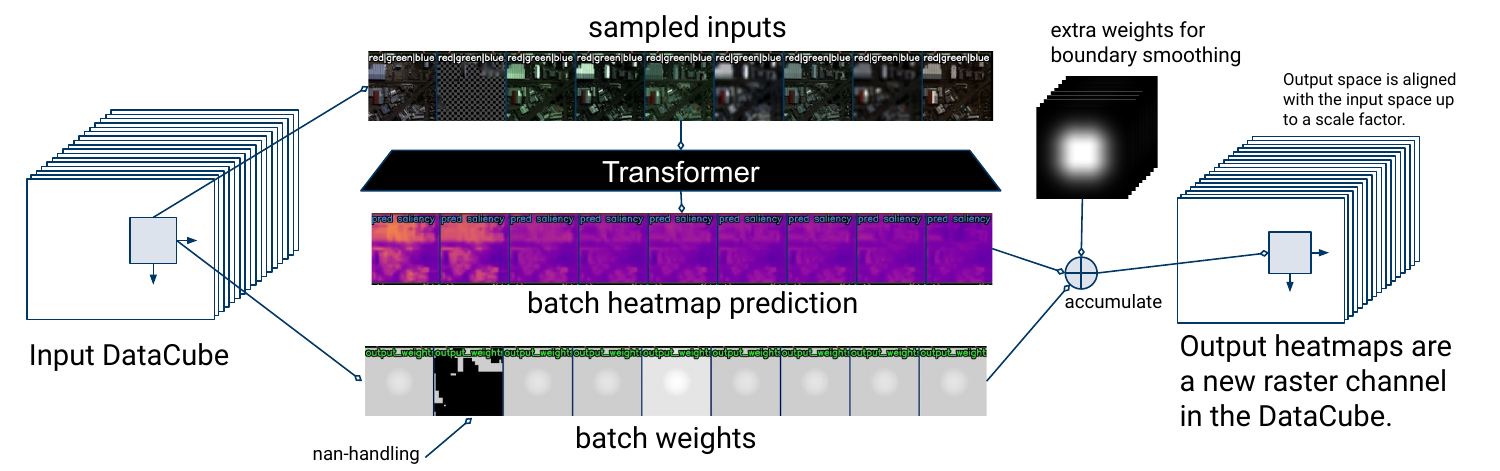}
    \captext[Prediction Pipeline.]{
    Given an enumeration of spacetime sample grids, the input is prepared and passed to a model, which predicts a corresponding set of heatmaps.
    With the input is an associated set of weights for each pixel, which is zero if the pixel is NaN or low quality as indicated by the QA mask.
    The heatmap predictions are accumulated into a pre-allocated buffer for for each frame in the larger video.
    Boundary smoothing weights are used to down-weight edges of each predicted window.
    When combined with overlapping windows, this results in a smooth final heatmap corresponding to each larger frame in the original video.
    }
    \label{fig:fusion_predict}
\end{figure*}

\section{Training and Inference}
\label{sec:approach:fusion} 

Given a dataset in KWCoco format, a network is trained by specifying an input window and resolution, a time kernel, sensor/channel specification that will be used to construct batches.
At creation time, the dataset estimates dataset statistics like mean/std and class frequency and cached.
The network also establishes trainable classes, mean/std, and class frequency by receiving them from the dataset.
For regularization we have implemented a restricted variant of the generate-and-test algorithm for continual learning \cite{dohare_loss_2023} as well as the full shrink-and-perturb \cite{ash_warm-starting_2020} algorithm. 
Our networks states are saved as checkpoints, and packaged with \texttt{torch.package}, which allows us to bundle the weights with the network topology and other metadata like train-time parameters.
All training details are logged in the the package metadata. 
For network details see previously published work \cite{greenwell_watch_2024}.


At test time, we construct a regular grid of sample targets, and produce an output for each input location. 
These outputs are stitched together in pre-allocated memory corresponding to the spacetime extent of the input dataset. 
The stitcher maintains an accumulation array, allowing overlapping windows to make multiple predictions for the same location in space time. 
The final pixel value at a location is the weighted average of all inputs accumulated there. 
Each prediction window augments the weights provided by the dataset (which indicates NaN locations) with  an additional filter that down-weights spatial edges and smooths boundary artifacts produced by overlapping windows. 
The prediction process is illustrated in \Cref{fig:fusion_predict}.

The result of this prediction process is either a new 1 channel saliency map, or a new per-class heatmap. 
The outputs are quantized to int16 and written to disk as geo-registered COGs and registered as new bands in the output KWCoco file.

\section{Partial Weight Initialization}
\label{sec:approach:partial-weight-loading}


\begin{figure}[t!]
    \centering
    \includegraphics[width=0.9\linewidth]{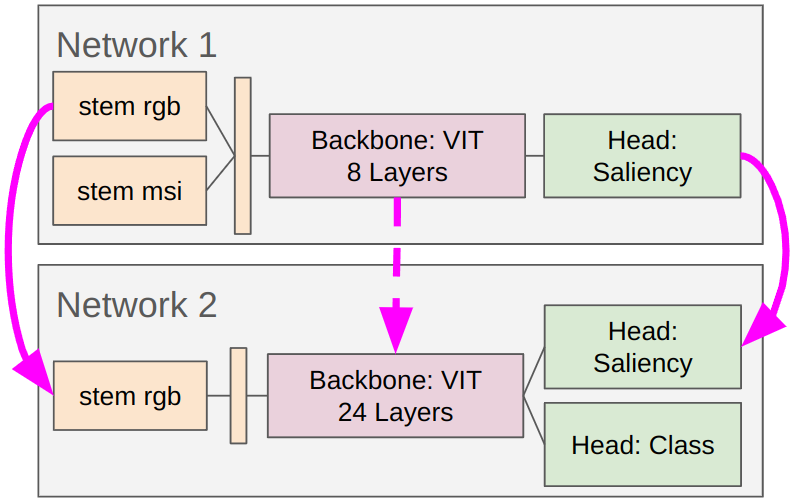}

    \captext[Partial Weight Loading.]{A partial matching between similar networks is established by finding a maximum common subtree embedding  \cite{feruglio_maximum_2003}. Unmatched destination weights are reinitialized. In this example the input stem MSI data is dropped, the backbone is extended from 8 layers to 24 layers and partially initialized, a new class head is initialized, and the existing saliency head and RGB stem are exactly copied.
    }
    \label{fig:matching}
\end{figure}


Our training pipeline has a unique initialization process. The network is either initialized from scratch by default or, if specified, weights are transferred using partial-weight-initialization \cite{torch_liberator}. The establishment of a partial matching between similar networks involves finding a maximum subgraph isomorphism \cite{feruglio_maximum_2003}, as shown in a simplified version in \Cref{fig:matching}.


Our networks possess extensive training history, with each network having a lineage of initialization events. This lineage comprises a chain of training events facilitated by partial weight loading, representing the model's training history and initialization states. We have leveraged this tool to evolve our architectures by adjusting depth, re-configuring input stems for different modalities, and modifying and re-initializing network heads for various objectives. Notably, training on semantically rich input features, such as activations of a pretrained MAE, enhances model performance. Interestingly, even after removing this feature and continuing training with RGB inputs, the model retains its improved performance --- a phenomenon we term "instillation."

\section{Detecting Heavy Construction Events}
\label{sec:approach:bas}

\begin{figure}
    \centering
    \includegraphics[width=1\linewidth]{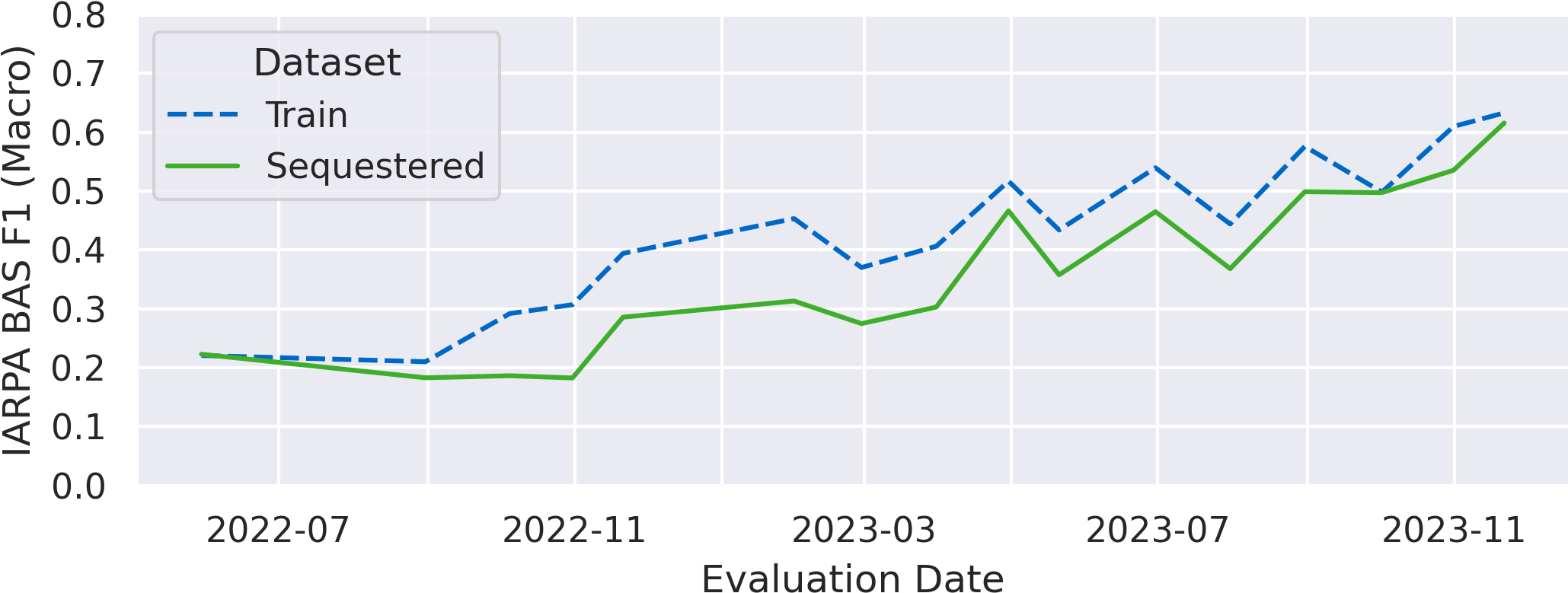}
    \captext[Instillation Improved Scoring Over Time.]{
    Over an 18-month period, our F1 scores for the IARPA SMART BAS task ($\uparrow$ is better) improved on both the training regions (dashed-blue) and the sequestered test regions (solid-green).
    Each model is finetuned under different training conditions from its most recent ancestor, initializing each step using our partial weight loading approach.
    The scores are reported to us from an external evaluation of our system.
    }
    \label{fig:phase2_scores}
\end{figure}

The motivating application of our system is the detection and classification of heavy construction events for the IARPA SMART challenge. Using the techniques described in this paper we continuously trained models that improved over time as shown in \Cref{fig:phase2_scores}. Our system works in two phases: broad-area-search (BAS) and activity characterization (AC). In BAS we search for candidate regions using data sampled at 10m GSD. In AC we zoom into candidate detections at 2m GSD, refine spacetime boundaries, and classify the phase of construction. 

\paragraph{Broad Area Search} For BAS our best network was trained with a combination of red, green, blue, and near-infrared bands from Landsat, Sentinel, and Worldview as well as a 36-band derived feature using the COLD algorithm \cite{ZHU2020111116}. 
We found that an important step was time-averaging (using the median) our input KWCoco files such that all images per-sensor per-year were averaged together to produce 1 image per sensor per year. An important note is that COLD features are only produced once per-year for Landsat 8 and Sentinel 2 data. 
Instead of averaging these features together we simply assign them to the nearest average image in time.

Given the trained BAS saliency model and a dataset, we predict saliency heatmaps for each frame.
We then extract the polygons from these heatmaps, which will become our initial predicted site boundaries. 
First, we find the maximum saliency response over all time. 
We use a threshold (0.375) to binarize this max-image, and produce polygons.
These are our BAS spatial bounds. 
Any polygon outside an area threshold of under $7200 m^2$ or above $8 km^2$ is removed.
We then assign temporal extents by looking again at the heatmaps. 
For each 1 year window (which may contain multiple predictions for different sensors) we average predictions together. 
For each of these frames, average saliency response under the polygon and assign it as the "score" for that spacetime observation. 
Any observation with a score under $0.3$ (chosen via grid search) is removed, which defines the start / end time for the site proposal. 

\paragraph{Activity Characterization} 
Given a set of site proposals, we cluster them into a set of smaller regions-of-interest. 
We build a KWCoco dataset by pulling Sentinel 2 and Worldview data with a maximum resolution of $2m$ GSD, downsampling all high-resolution sources. 
We pull all available Worldview imagery, but we consume only least cloudy Sentinel 2 image per month.

Predicting class heatmaps works almost exactly as in BAS. 
For each video we define a target grid, push each input through the network and 
we output a new dataset with \emph{both} 4-channel class heatmaps and high resolution saliency heatmaps. 

To extract high-resolution polygons, we employ a tracking process for each site cluster. We load predicted heatmaps for each cluster, multiply the "Active Construction" and "ac-salient" channels, and construct a volume. Pixels inside each BAS polygon are considered, and all other pixels are zeroed. The tensor is binarized based on a threshold (e.g., 0.3), and connected components are found. The maximum response in each component serves as seed points for a watershed algorithm, filling the BAS polygon area with smaller polygons. Small polygons are removed. The maximum AC-salient score is used as a singular filter for each site. A site is rejected if its singular score is below 0.3, and if the maximum per-class score is not above 0.3, the observation is labeled as "No Activity." The start date for a site is determined by the first "Site Preparation" observation, and the end date is determined by the last "Active Construction" observation. Any site without "Site Preparation" or "Active Construction" predictions is rejected.

\section{Discussion}
\label{sec:discussion}

We have introduced GeoWATCH, a software framework for training and predicting AI models on heterogeneous raster time sequences. The system handles tasks like image classification, activity recognition, object detection, or object tracking based on MS-COCO-compatible input data encoding. We demonstrated its effectiveness in detecting heavy construction events in diverse satellite image sequences. Our training system features a novel partial weight loading mechanism, utilizing sub-graph isomorphism, enabling continuous network training and modification over extended periods. This facilitates the creation of a lineage of models, showing improved performance over time by adjusting configurations while maintaining a core backbone. Notably, training models with many features and initializing new networks with fewer input features retains performance, suggesting an "instillation" of knowledge from previous models into new ones. Future research will explore experimental verification of these observations.

 {\small \textbf{Acknowledgement}
This research is based upon work supported in part by the Office of the
Director of National Intelligence (ODNI), 6 Intelligence Advanced Research
Projects Activity (IARPA), via 2021-2011000005. The views and conclusions
contained herein are those of the authors and should not be interpreted as
necessarily representing the official policies, either expressed or implied, of
ODNI, IARPA, or the U.S. Government. The U.S. Government is authorized to
reproduce and distribute reprints for governmental purposes notwithstanding any
copyright annotation therein.
}

\bibliographystyle{IEEEbib}
\bibliography{igaars-main}

\end{document}